\documentclass{article}

\usepackage[preprint]{neurips_2026}

\usepackage[utf8]{inputenc}%
\usepackage[T1]{fontenc}%
\usepackage{hyperref}%
\usepackage{url}%
\usepackage{booktabs}%
\usepackage{amsfonts}%
\usepackage{nicefrac}%
\usepackage{microtype}%
\usepackage{xcolor}%

\usepackage{graphicx}
\usepackage{subcaption}
\usepackage{multirow}
\usepackage{algorithm}
\usepackage{algorithmic}
\usepackage{amsmath}
\usepackage{amssymb}
\usepackage{mathtools}
\usepackage{amsthm}

\usepackage[capitalize,noabbrev]{cleveref}

\theoremstyle{plain}
\newtheorem{theorem}{Theorem}[section]

\newtheorem{lemma}[theorem]{Lemma}

\theoremstyle{definition}

\theoremstyle{remark}

\newsavebox{\mainresultsfigbox}
\newsavebox{\mainresultstabbox}
\newlength{\mainresultsheight}

\title{Is the Importance Ratio Necessary for Stable Reinforcement Learning in LLMs?}

\author{%
  Shuibai Zhang\thanks{Equal contribution.}\hspace{0.45em}\thanks{Work done during an internship at KRAFTON.}\hspace{1em} \\
  University of \\
  Wisconsin--Madison \\
  \And
  Junhyuck Kim\footnotemark[1]\hspace{0.45em} \\
  KRAFTON \\
  \And
  Gyeongman Kim \\
  KRAFTON \\
  \And
  Jaewoong Cho \\
  KRAFTON \\
}

\begin{document}

\maketitle

\begin{abstract}
  Reinforcement learning (RL) has become central to post-training large language models (LLMs).
  However, popular RL methods like GRPO incur non-negligible overhead by computing both old-policy and current-policy likelihoods to form importance sampling ratios.
  In this work, we propose Likelihood-Gated Policy Optimization (LGPO), which enforces a soft trust region constraint via likelihood-based gating, eliminating the need to compute old-policy likelihoods.
  Empirically, we show that removing the importance sampling correction term does not harm training stability, whereas removing the trust region mechanism can lead to collapse.
  Moreover, ratio-based clipping can fail in fully on-policy training: the importance ratio stays at 1, so the ratio-based trust region constraint never activates.
  Under standard training settings where GRPO is stable, LGPO achieves comparable training stability and peak performance while reducing training time by $\sim$18\% on average relative to a standard implementation that recomputes old-policy likelihoods.
  In fully on-policy training, where GRPO fails, LGPO remains stable, enabling more efficient and robust LLM RL post-training across training regimes.
\end{abstract}

\section{Introduction}

Reinforcement learning (RL) plays a significant role in the post-training of modern large language models (LLMs)~\citep{jaech2024openai, guo2025deepseek}.
In particular, our focus is on RL with verifiable rewards, aiming to enhance the reasoning capabilities of LLMs~\citep{zhang2025survey}.
A central challenge in this field is achieving stable RL training while maintaining computational efficiency.
To accelerate training in standard on-policy settings, it is common practice to perform multiple parameter updates per rollout batch.
However, this introduces a distribution shift between the current policy and the data-generating (old) policy.

To address this deviation, importance sampling ratios, computed as the ratio of the current policy likelihood to the old policy likelihood, are typically used to correct the distribution mismatch and reduce gradient estimate bias.
Complementary to this, trust region methods are employed to explicitly ensure stability by constraining the extent of policy updates.
While TRPO~\citep{schulman2015trust} enforces an explicit KL divergence constraint, modern algorithms like PPO~\citep{schulman2017proximal} and GRPO~\citep{shao2024deepseekmath} utilize importance ratio-based clipping to impose a soft trust region.

Despite being crucial for advancing LLM capabilities, RL incurs high computational cost, particularly for reasoning tasks that involve generating and learning from long sequences.
Although recent works have sought to improve training and sample efficiency~\citep{yu2025dapo,yue2025vapo,chen2025minimax,yao2025flashrl,zheng2025act,zheng2025group}, they retain the reliance on importance sampling ratios to correct for distribution shift.
This imposes a non-negligible overhead: for example, obtaining probabilities of 128 sequences of length 32K tokens for a 32B model requires approximately $4.06 \times 10^{17}$ FLOPs.
We challenge this standard practice, questioning whether the computational cost of importance sampling is strictly necessary for stability when policy drift is otherwise managed.
In this work, we therefore ask:

\begin{center}
\textit{Can we achieve more efficient LLM RL training without having to compute importance sampling ratios and maintain training stability?}
\end{center}
\vspace{-0.05cm}

In Section~\ref{sec:removing_importance_weight}, we demonstrate that the importance sampling correction is not essential for stability, whereas the trust region mechanism is critical.
Moreover, as we show in Section~\ref{sec:fully_on_policy}, ratio-based clipping fails in the fully on-policy setting, where a single policy update is performed per rollout batch: the importance ratio remains constant at 1, effectively removing the trust region constraint and leading to collapse.

Based on these observations, we propose Likelihood-Gated Policy Optimization (LGPO) in Section~\ref{sec:trust_region_via_likelihood_based_gating}, along with a theoretical analysis motivating the use of likelihood-based gating.
LGPO enforces a trust region via likelihood-based gating without computing importance ratios.
Experiments on Qwen2.5 models~\citep{qwen2025qwen25technicalreport,yang2024qwen2} across Countdown and mathematical reasoning tasks show that LGPO matches GRPO's stability and performance while reducing training time by $\sim$18\% relative to the standard recipe that recomputes old-policy likelihoods (Figure~\ref{fig:main_training_curves} and Table~\ref{tab:main_results}).
Furthermore, LGPO remains stable in the fully on-policy setting where GRPO fails (Figure~\ref{fig:fully_on_policy_lgpo_vs_grpo}), even when a KL penalty is added to GRPO (Section~\ref{sec:main_results}).
We further perform case studies and ablations to better understand LGPO and the mechanism behind its stability (Section~\ref{sec:ablation_studies_and_analysis}).

Overall, our contributions are summarized as follows:
\begin{enumerate}
    \item We dissect the role of importance ratios, showing that in common LLM RL settings their stabilizing effect comes from the trust region implemented by clipping, while the correction term is dispensable across GRPO, Dr.~GRPO, DAPO, and GSPO.
    \item We identify a failure mode of ratio-based clipping in fully on-policy training, where the importance ratio is identically $1$ and clipping never constrains updates.
    \item We propose LGPO, which enforces a trust region via likelihood-based gating using only the current policy's likelihood, without old- or reference-policy log-probabilities. LGPO matches GRPO with less training time and remains stable in the fully on-policy setting where GRPO, with or without a small KL penalty, collapses.
\end{enumerate}

\section{Background}

A vanilla policy gradient algorithm in the context of RL training of LLMs aims to maximize the following objective:

\begin{equation*}
  \mathcal{L}^{\text{PG}}(\theta) = \mathbb{E}_{x \sim \mathcal{D}, y \sim \pi_\theta(\cdot|x)} \left[ \sum_{t=1}^{T} A_t \log \pi_\theta(y_t|x, y_{<t}) \right],
\end{equation*}

where $x$ is a prompt sampled from the prompt distribution $\mathcal{D}$, and $y=(y_1,\ldots,y_T)$ is a sampled response of length $T$ from the policy $\pi_\theta(\cdot\mid x)$.
$y_t$ denotes the token at position $t$ and $y_{<t}$ its prefix.
$A_t$ is the advantage for token $t$.

REINFORCE~\citep{williams1992simple} is a specific implementation of the vanilla policy gradient that uses a Monte Carlo estimator of the gradient, where the sampling is performed \textit{fully on-policy}, which we define as the case where the data is always sampled from the current policy that is being optimized:

\begin{equation*}
  \nabla_\theta \mathcal{L}^{\text{PG}}(\theta) \approx \frac{1}{N} \sum_{i=1}^{N} \sum_{t=1}^{T_i} A_{i,t} \nabla_\theta \log \pi_\theta(y_{i,t}|x_i, y_{i,<t}).
\end{equation*}
Here $N$ is the batch size, and $A_{i,t}$ is the advantage for the $t$-th token of the $i$-th sample.

Trust Region Policy Optimization (TRPO)~\citep{schulman2015trust} introduces the setting where multiple parameter updates using the samples generated from the ``old policy'' $\pi_{\theta_\text{old}}$ are possible, deviating from the fully on-policy setting.
It is now common practice in LLM RL training to perform multiple parameter updates in a mini-batch manner to accelerate the training process.
During these multiple parameter updates, a hard constraint on the KL divergence from the old policy is applied, preventing large policy changes, thus stabilizing the training.
The gradient estimate is now a Monte Carlo estimate with \textit{importance sampling ratio} correcting for the distribution mismatch between the old policy and the current policy.
The TRPO objective is defined as:

\begin{equation*}
  \mathbb{E}_{x \sim \mathcal{D}, y \sim \pi_{\theta_\text{old}}(\cdot|x)} \left[ \sum_{t=1}^{T} r_t(\theta) A_t \right] \quad \text{s.t.}\quad \mathbb{E} \left[ \sum_{t=1}^{T} \mathrm{KL}\big(\pi_{\theta_\text{old}}(\cdot|x, y_{<t}) \,\|\, \pi_\theta(\cdot|x, y_{<t})\big) \right] \le \delta.
\end{equation*}

Here $r_t(\theta) = \frac{\pi_\theta(y_t|x, y_{<t})}{\pi_{\theta_\text{old}}(y_t|x, y_{<t})}$ is the importance sampling ratio and $\delta$ is a hyperparameter bounding the KL divergence.
The expectation in the constraint is taken over $x \sim \mathcal{D}$ and $y \sim \pi_{\theta_\text{old}}(\cdot|x)$.

Proximal Policy Optimization (PPO)~\citep{schulman2017proximal} replaces the hard KL constraint of TRPO with the clipping operation shown in Eq.~\ref{eq:ppo}, which removes the incentive for moving the importance ratio outside of the interval $[1-\epsilon, 1+\epsilon]$, where $\epsilon$ is a hyperparameter.
The PPO objective is defined as:

\begin{equation}
  \label{eq:ppo}
  \mathbb{E}_{x \sim \mathcal{D},\, y \sim \pi_{\theta_\text{old}}(\cdot|x)} \left[ \sum_{t=1}^{T} \min(r_t(\theta) A_t, \operatorname{clip}(r_t(\theta), 1-\epsilon, 1+\epsilon) A_t) \right].
\end{equation}

This choice of clipping the objective based on the importance ratio can be viewed as serving two intertwined purposes: (1) discouraging updates that would push the policy far from the old policy, creating a ``soft trust-region'' effect, and (2) avoiding the use of samples with high importance ratios, thus reducing the variance of the gradient estimate.

Group Relative Policy Optimization (GRPO)~\citep{shao2024deepseekmath} is a variant of PPO that offers better efficiency and scalability.
Instead of computing the advantage $A_t$ using a learned value function via GAE~\citep{schulman2015high}, GRPO eliminates the value model by sampling a group of $G$ responses $\{y_1, \dots, y_G\}$ for each prompt $x$ and computing advantages based on the relative rewards within the group.
Specifically, for the $i$-th response in the group with reward $R_i$, the advantage is given by:
\begin{equation*}
  A_i = \frac{R_i - \mu_{\text{group}}}{\sigma_{\text{group}}},
\end{equation*}
where $\mu_{\text{group}}$ and $\sigma_{\text{group}}$ are the mean and standard deviation of the rewards in the group.
GRPO then optimizes the PPO objective (Eq.~\ref{eq:ppo}) using these group-relative advantages.
Note that we do not incorporate a KL divergence penalty relative to the initial reference policy in this work, as training for complex reasoning tasks often requires significant deviation from the initialization~\citep{yu2025dapo}; we evaluate a KL-regularized variant as a baseline in Section~\ref{sec:main_results}.

Following PPO, modern RL algorithms specialized for LLM post-training, including GRPO and its variants~\citep{zheng2025group,yu2025dapo,yue2025vapo,chen2025minimax}, use the importance ratio to correct for distribution mismatch and implement clipping.
However, computing the importance ratio incurs additional memory and computational overhead at each sampling step.
Given that in many practical LLM RL training settings the number of parameter updates per sampling step is small (e.g., 4), we ask: \textit{is it possible to remove the importance ratio to achieve more efficient yet stable training?}

\section{Are Importance Ratios Necessary for Stable Training?}

\subsection{Importance Sampling Correction is Dispensable}
\label{sec:removing_importance_weight}

In PPO/GRPO, the importance ratio $r_t(\theta) = \frac{\pi_\theta(y_t|x, y_{<t})}{\pi_{\theta_\text{old}}(y_t|x, y_{<t})}$ plays two distinct roles.
We can explicitly see this by examining the gradient of the surrogate objective for a single token at $t$ (see derivation in Appendix~\ref{app:gradient_derivation}):
\begin{equation}
\label{eq:grad_decomposition}
\nabla_\theta L_t(\theta) = \underbrace{\mathbf{1}_{\text{clip}}}_{\text{clipping}} \cdot A_t \cdot \underbrace{r_t(\theta)}_{\text{correction}} \cdot \nabla_\theta \log \pi_\theta(y_t|x, y_{<t}),
\end{equation}
where $\mathbf{1}_{\text{clip}}$ is an indicator function that is 1 when the ratio is within the clipping range and 0 otherwise (see Appendix~\ref{app:gradient_derivation} for the full conditions).
The term $r_t(\theta)$ in Equation~\ref{eq:grad_decomposition} acts as a \textbf{correction} weight accounting for distribution shift, while $\mathbf{1}_{\text{clip}}$ implements \textbf{clipping}, the soft trust region constraint.
The correction weight makes the surrogate an unbiased estimate of the on-policy objective when training on samples from $\pi_{\theta_\text{old}}$, so removing it biases the gradient (Appendix~\ref{app:variance_from_large_importance_ratios}).
Our claim is thus empirical rather than statistical: the clipped surrogate is already biased, and we ask whether the additional bias matters for stability when $N_\text{updates}$ is small, as in common LLM RL settings.

In this section, we examine the effect of removing each of these two roles one by one, and show that removing the correction role is not harmful to training stability in common LLM RL training settings, but removing the clipping role causes training instability.

We experiment with RL training of Qwen2.5-3B-Instruct on the Countdown task.\footnote{Countdown is a verifiable numerical task where, given a set of numbers, the model must combine them using basic arithmetic operations to reach a target number.}
We use GRPO as our baseline algorithm.
First, we remove the correction role by setting the gradient scaling term to 1, while preserving the ratio-based clipping mechanism (\textit{w/o correction}).
Second, we remove the clipping mechanism while retaining the importance sampling correction (\textit{w/o clipping}).
We denote the number of gradient updates performed per batch of sampled data as $N_\text{updates}$ and use $N_\text{updates}=2$ for this experiment.

\begin{figure}[t!]
  \centering
  \includegraphics[width=0.5\linewidth]{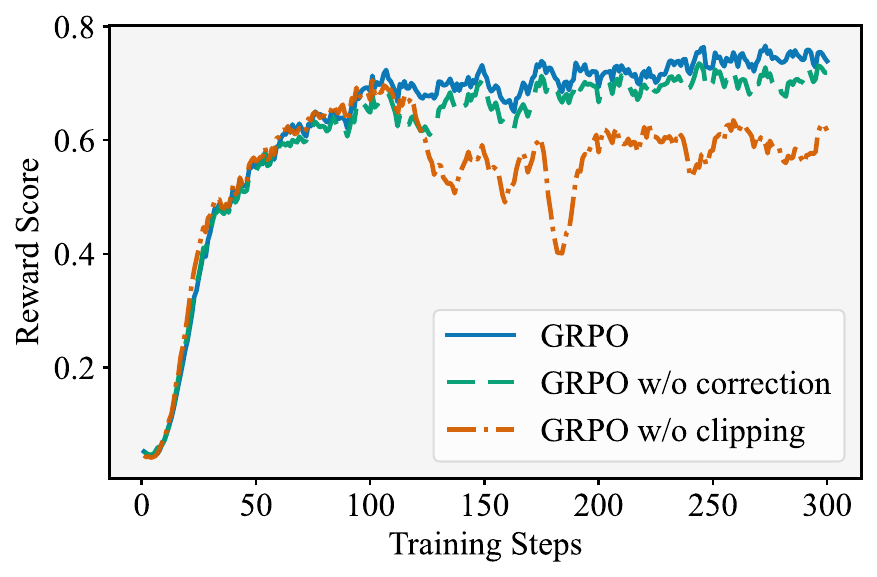}
  \caption{
    The reward score of the Countdown task during the RL training.
    We compare removing the importance sampling \textbf{correction} ($r_t(\theta)$ from Eq.~\ref{eq:grad_decomposition}) versus removing ratio-based \textbf{clipping} ($\mathbf{1}_{\text{clip}}$ from Eq.~\ref{eq:grad_decomposition}).
    Removing the correction has little effect, while removing clipping leads to significant instability.
    Runs in this section are trained for 300 steps, sufficient to show the stability contrast.
  }
  \label{fig:removing_importance_weight}
\end{figure}

We find that removing only the importance sampling correction from GRPO (\textit{w/o correction}) has a negligible effect on training stability and only marginally worsens the scaling behavior, as illustrated in Figure \ref{fig:removing_importance_weight}.
On the other hand, removing the clipping mechanism (\textit{w/o clipping}) leads to significant instability.
We observe similar findings for other GRPO-variants including DAPO~\citep{yu2025dapo} and GSPO~\citep{zheng2025group}: removing the correction term alone is not harmful to training stability (see Appendix~\ref{app:dapo_gspo_removing_correction} and Figure~\ref{fig:dapo_gspo_removing_correction}).
The same holds for Dr.~GRPO~\citep{liu2025understanding}, which does not divide advantages by the group standard deviation; there, removing clipping also did not destabilize training within 300 steps, which we attribute to its smaller per-sample updates (Appendix~\ref{app:dr_grpo}).
In contrast, a GPG-style objective~\citep{chu2026gpg} with neither importance ratio nor clipping collapses in our setup unless a sufficiently strong KL penalty is added (Appendix~\ref{app:gpg}).
Training without any trust region can thus be stable in some configurations, but not uniformly so.

These results indicate that the importance sampling correction is not strictly necessary, provided that a trust region constraint is maintained.
Recall that importance ratio-based clipping serves two purposes: (1) enforcing a soft trust region, and (2) reducing variance from large importance ratios.
With the correction term removed, the variance amplification caused by large importance ratios is eliminated, leaving clipping solely to enforce the trust region (see Appendix~\ref{app:variance_from_large_importance_ratios} for variance analysis).
However, as we show next, ratio-based clipping fails to provide any constraint in the fully on-policy setting, motivating a trust region mechanism independent of the importance ratio.

\subsection{Importance Ratio-Based Clipping Fails in Fully On-Policy Training}
\label{sec:fully_on_policy}

Recall that $N_\text{updates}$ is the number of gradient updates performed per batch of sampled data.
We define the \textit{fully on-policy} setting as the case where $N_\text{updates}=1$.
In this setting, we sample rollouts using the current policy $\pi_\theta$ and perform exactly one gradient update using this data.
Consequently, the behavior (old) policy $\pi_{\theta_{\text{old}}}$ is always identical to the current policy $\pi_\theta$, and the importance ratio $r_t(\theta)$ remains fixed at 1.
Since clipping is triggered only when the ratio deviates from 1, it is never activated in this setting, failing to serve as a trust region constraint.

\begin{figure}[t!]
  \centering
  \includegraphics[width=\linewidth]{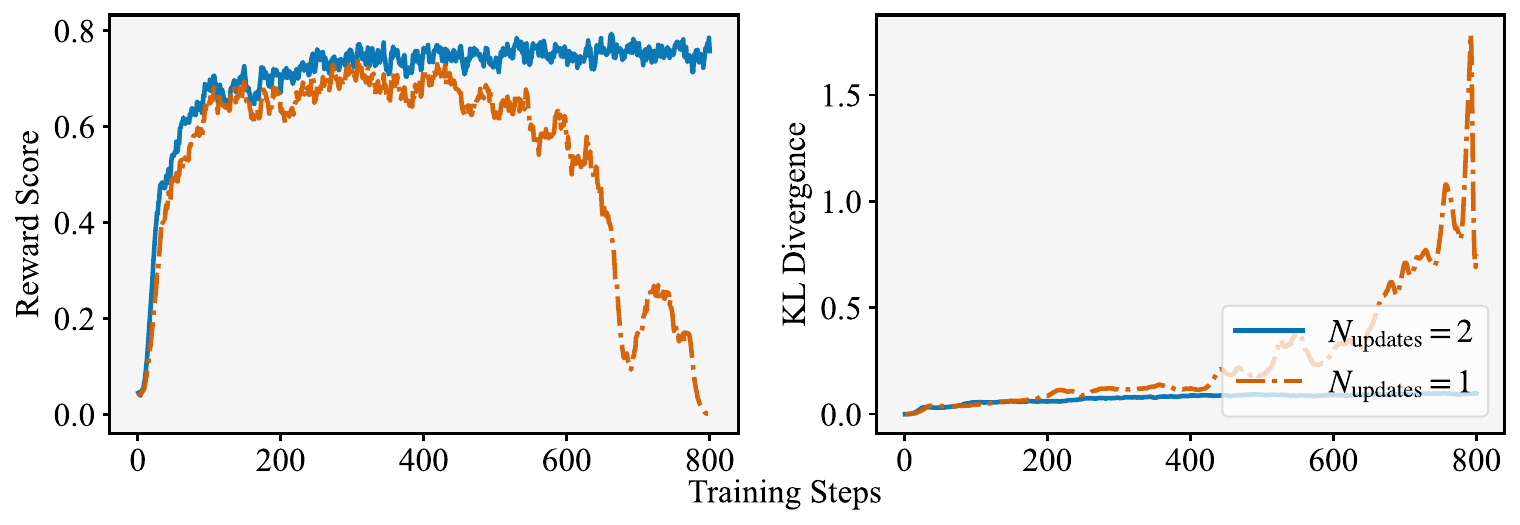}
  \caption{
    Performance comparison of GRPO with $N_\text{updates}=1$ (fully on-policy) and $N_\text{updates}=2$ on the Countdown task.
    $N_\text{updates}$ denotes the number of gradient updates performed per batch of sampled data.
    We show the reward score (top) and the KL divergence from the reference policy (bottom).
    Without a functioning trust region constraint ($N_\text{updates}=1$), training collapses and the policy diverges significantly from the reference model.
    Fully on-policy runs are trained for 800 steps because collapse in this regime emerges later in training.
  }
  \label{fig:fully_on_policy}
\end{figure}

We empirically observe that this leads to training collapse under the same experimental settings as in Section \ref{sec:removing_importance_weight}.
As shown in Figure \ref{fig:fully_on_policy}, fully on-policy GRPO, which effectively has no trust region constraint, collapses after an initial phase of improvement.
We also observe that the policy drifts far away from the initial reference policy.

Since the importance ratio is thus only required for the trust region constraint (Section~\ref{sec:removing_importance_weight}), relying on it has two drawbacks: (1) the overhead of recomputing old-policy likelihoods with the training engine (Section~\ref{sec:main_results}), and (2) no constraint at all in the fully on-policy setting.
This motivates a trust region mechanism that is both efficient and robust across training regimes.

\section{Trust Region via Likelihood-Based Gating}
\label{sec:trust_region_via_likelihood_based_gating}

We propose a simple alternative soft trust region constraint: likelihood-based gating, where the likelihood is the probability of the \textit{current} policy generating the sampled data, so no old policy probabilities need to be computed or stored.
To motivate this, we first analyze how the sample likelihood relates to the sensitivity of the policy update (measured by KL divergence) induced by the sample.

\begin{theorem}[Minimal KL divergence under a single-outcome perturbation]
  \label{thm:local-kl-sensitivity}
  Let $p\in\Delta^{K-1}$ be a categorical distribution over a finite outcome space $\{1,\ldots,K\}$, with $p_i>0$.
  Consider another distribution $q=p+\Delta p$, where $\sum_{i=1}^K \Delta p_i=0$ and $q_i>0$.

  Fix an outcome $k$ and enforce $\Delta p_k=\delta$ with $-p_k<\delta<1-p_k$. Then, as $|\delta|\to 0$,
  \[
  \inf_{\substack{\Delta p_k=\delta\\ \sum_i \Delta p_i=0}}
  \mathrm{KL}\big(p(\cdot) \,\|\, q(\cdot)\big)
  =
  \frac{\delta^2}{2\,p_k(1-p_k)} + o(\delta^2).
  \]
  In particular, for a fixed probability change $|\delta|$, the induced KL change grows as $p_k\to 0$ or $p_k\to 1$.
\end{theorem}
The proof is provided in Appendix~\ref{app:proof-local-kl-sensitivity}.
Since the outcome space is arbitrary, the theorem applies both at the \emph{token level}, with $p=\pi(\cdot\mid x,y_{<t})$ the next-token distribution and $q=\pi'(\cdot\mid x,y_{<t})$ that of an updated policy $\pi'$, and at the \emph{sequence level}, with $p=\pi(\cdot\mid x)$ the distribution over responses, which is finite because responses have bounded length over a finite vocabulary.
In the context of LLM RL training, learning from a sample amounts to increasing or decreasing its likelihood under the policy.
Theorem~\ref{thm:local-kl-sensitivity} implies that for a fixed magnitude of likelihood change, the induced policy change (measured by KL divergence) is largest when the sample likelihood is close to 0 or 1.
This motivates our strategy of gating out samples with extreme likelihoods to mitigate these highly sensitive updates, thereby enforcing a soft trust region constraint.

\paragraph{Likelihood-Gated Policy Optimization (LGPO).}
Based on this motivation, we propose LGPO, which enforces the trust region constraint directly via likelihood-based gating.
Three considerations determine the gating statistic:
(i) Theorem~\ref{thm:local-kl-sensitivity} identifies extreme likelihood as a sensitivity signal;
(ii) with verifiable rewards, the advantage is assigned per sequence, so we gate at the sequence level to align the trust region with the unit of learning~\citep{zheng2025group}; and
(iii) averaging the token log-likelihoods (the log of their geometric mean) normalizes for length, so a single threshold is comparable across responses.
We thus define the \emph{sequence-level average log-likelihood} under the current policy $\pi_\theta$ as
\begin{equation*}
  \bar{\ell}_\theta(x,y)
  := \frac{1}{T}\sum_{t=1}^{T} \log \pi_\theta(y_t \mid x, y_{<t}).
\end{equation*}
A sequence with high $\bar{\ell}_\theta(x,y)$ is dominated by tokens from near-saturated next-token distributions ($p_k\to 1$), the regime that the token-level instantiation of Theorem~\ref{thm:local-kl-sensitivity} identifies as most sensitive.
We only gate high-likelihood sequences: for a sequence-level average, a low $\bar{\ell}_\theta(x,y)$ typically reflects many moderate-probability tokens, where the sensitivity $\delta^2/(2p_k(1-p_k))$ is small, and such sequences are rarely observed (Appendix~\ref{app:likelihood_distribution}); additionally gating them yields no benefit (Appendix~\ref{app:two_sided}).

We thus define the gating mask as
\begin{equation*}
  c_\theta(x,y) := \mathbf{1}(\bar{\ell}_\theta(x,y) \le \tau),
\end{equation*}
where $\tau$ is a likelihood threshold in log-likelihood space.
This yields the LGPO objective:
\begin{equation*}
  \mathcal{L}^{\text{LGPO}}(\theta)
  =
  \mathbb{E}_{x \sim \mathcal{D},\, y \sim \pi_{\theta_{\text{old}}}(\cdot\mid x)}
  \left[ c_\theta(x,y)\, \sum_{t=1}^{T} A_t \log \pi_\theta(y_t \mid x, y_{<t}) \right],
\end{equation*}
where $c_\theta(x,y)$ serves as a stop-gradient mask, effectively zeroing out the gradient contribution from samples that exceed the likelihood threshold.
The full training procedure can be found in Algorithm~\ref{alg:clpo} in Appendix~\ref{app:algorithmic-details}.

\begin{figure}[t!]
  \centering
  \includegraphics[width=\linewidth]{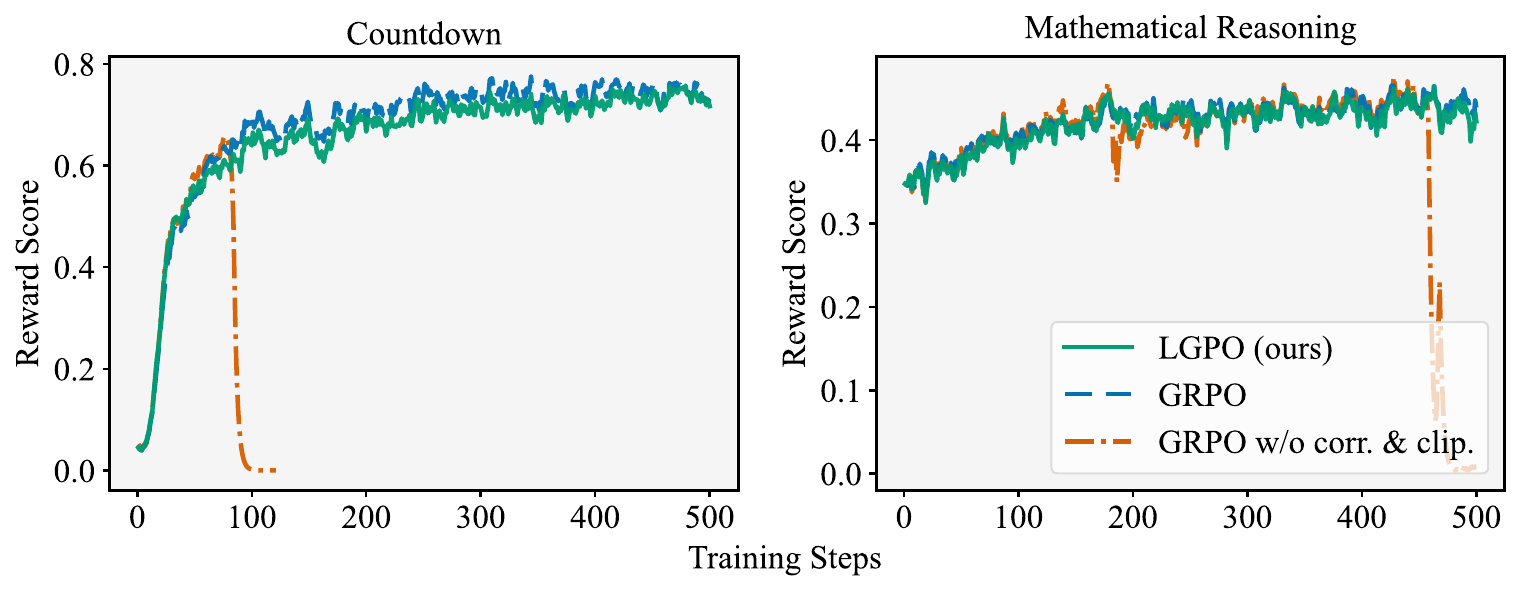}
  \caption{Training curves comparing LGPO, GRPO, and GRPO without importance sampling correction and clipping on Countdown (top) and mathematical reasoning (bottom), trained for 500 steps. LGPO maintains stability comparable to GRPO, indicating that likelihood-based gating serves as an effective alternative trust region constraint.}
  \label{fig:main_training_curves}
\end{figure}

\section{Experiments}
\label{sec:experiments}

We experiment with Qwen2.5-Math-1.5B and Qwen2.5-3B-Instruct.
We first analyze training dynamics on the Countdown task, then extend to mathematical reasoning and report final performance on MATH500~\citep{lightman2024let}, OlympiadBench~\citep{he2024olympiadbench}, and AIME24\&25~\citep{aime2025problems} (average Pass@1 over 8 samples), along with training curves.
For mathematical reasoning, we train on the high-quality dataset curated by \citet{huan2025does}.\footnote{\url{https://huggingface.co/datasets/ReasoningTransferability/math_rl_48k}}
Unless otherwise specified, during RL training, we use a minibatch size of 64.
For Countdown, we use a prompt batch size of 128 and perform two gradient updates per rollout batch ($N_\text{updates}=2$).
For mathematical reasoning, we use a prompt batch size of 256 and perform four gradient updates per rollout batch ($N_\text{updates}=4$).
We compute advantages using group relative normalization as in GRPO with a group size of $G=8$.
For LGPO, we set the likelihood threshold $\tau=-0.15$, which corresponds to $\exp(\tau)\approx 0.86$ average likelihood in probability space.
In practice, we recommend setting $\tau$ near the top 10--20\% quantile of the sequence-level average likelihood of sampled responses (cf.\ Figure~\ref{fig:likelihood_distribution}); the same $\tau$ is used unchanged across both models and both tasks (sensitivity analysis in Appendix~\ref{app:tau_sensitivity}).
We compare the LGPO objective against GRPO and GRPO \textit{w/o correction \& clipping} (introduced in Section~\ref{sec:removing_importance_weight}).
All methods are implemented in verl~\citep{sheng2025hybridflow} with identical hyperparameters (Appendix~\ref{app:hyperparameters}); only the training horizon differs by purpose: 300 steps for the analyses in Section~\ref{sec:removing_importance_weight}, 500 steps for the main comparisons, and 800 steps for fully on-policy runs, where collapse emerges later.

\subsection{Main Results}
\label{sec:main_results}

\begin{figure*}[t!]
  \centering
  \sbox{\mainresultsfigbox}{%
    \begin{minipage}[t]{0.52\textwidth}
      \vspace{0pt}
      \centering
      \includegraphics[width=\linewidth]{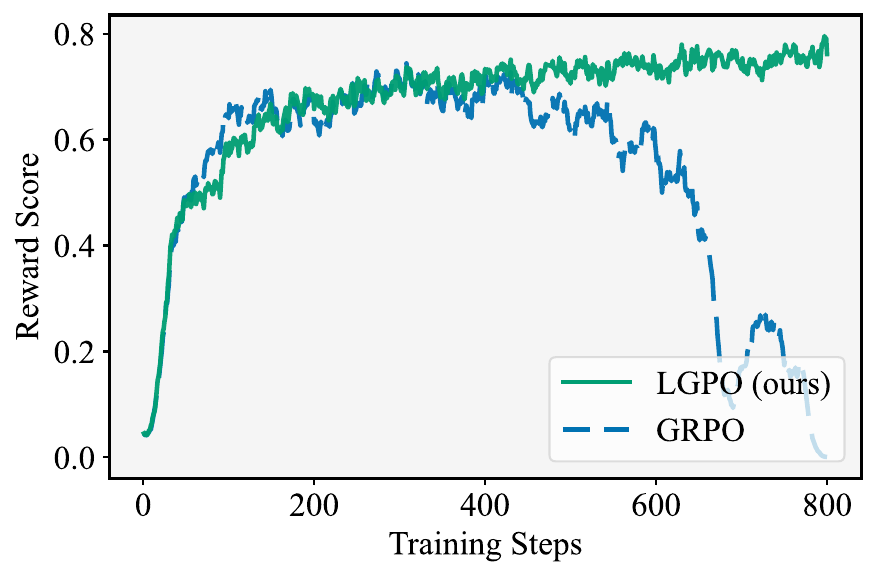}
      \caption{
        Fully on-policy ($N_\text{updates}=1$) training dynamics on Countdown using Qwen2.5-3B-Instruct (800 steps).
        GRPO collapses due to the lack of an effective trust region constraint when the importance ratio is fixed at $1$, whereas LGPO remains stable via likelihood-based gating.
      }
      \label{fig:fully_on_policy_lgpo_vs_grpo}
    \end{minipage}%
  }%
  \sbox{\mainresultstabbox}{%
    \begin{minipage}[t]{0.36\textwidth}
      \vspace{0pt}
      \centering
      \small
      \captionof{table}{Final evaluation performance on mathematical reasoning benchmarks.}
      \label{tab:main_results}
      \vspace{-2pt}
      \begin{tabular}{lcc}
        \toprule
        Benchmark & GRPO & LGPO \\
        \midrule
        \multicolumn{3}{c}{\textit{Qwen2.5-3B-Instruct}} \\
        MATH500 & 64.4 & \textbf{65.2} \\
        OlympiadBench & 29.2 & \textbf{30.7} \\
        AIME24 & \textbf{5.0} & 4.2 \\
        AIME25 & 3.3 & \textbf{3.7} \\
        \textit{Avg.} & \textit{25.5} & \textbf{\textit{26.0}} \\
        \midrule
        \multicolumn{3}{c}{\textit{Qwen2.5-Math-1.5B}} \\
        MATH500 & 73.0 & \textbf{73.6} \\
        OlympiadBench & \textbf{38.2} & 35.0 \\
        AIME24 & 12.1 & \textbf{17.9} \\
        AIME25 & 8.8 & \textbf{9.6} \\
        \textit{Avg.} & \textit{33.0} & \textbf{\textit{34.0}} \\
        \bottomrule
      \end{tabular}
      \vspace{-6pt}
    \end{minipage}%
  }%

  \setlength{\mainresultsheight}{\dimexpr\ht\mainresultstabbox+\dp\mainresultstabbox\relax}
  \ifdim\dimexpr\ht\mainresultsfigbox+\dp\mainresultsfigbox\relax>\mainresultsheight
    \setlength{\mainresultsheight}{\dimexpr\ht\mainresultsfigbox+\dp\mainresultsfigbox\relax}
  \fi

  \begin{minipage}[c][\mainresultsheight][c]{0.4\textwidth}
    \centering
    \usebox{\mainresultstabbox}
  \end{minipage}\hfill
  \begin{minipage}[c][\mainresultsheight][c]{0.56\textwidth}
    \centering
    \usebox{\mainresultsfigbox}
  \end{minipage}
\end{figure*}

Figure~\ref{fig:main_training_curves} shows that \textbf{LGPO achieves stable training without using importance ratios} on both Countdown and the mathematical reasoning training dataset.
While removing importance sampling correction and clipping from GRPO leads to unstable training (\textit{GRPO w/o corr. \& clip.}), adding simple likelihood-based gating, yielding the LGPO objective, stabilizes training.
We observe a negligible drop in peak reward score compared to GRPO.
Table~\ref{tab:main_results} reports evaluation scores after training for the same number of steps, demonstrating similar performance in learning mathematical reasoning capabilities that generalize to broader tasks.

While maintaining similar performance, \textbf{LGPO offers training speedups} by eliminating the need to compute old policy likelihoods.
When training Qwen2.5-3B-Instruct on $\sim$7.5K training data using 2 NVIDIA H100 GPUs, GRPO takes 3.37 hours to train one epoch, while LGPO takes 2.85 hours, corresponding to a $\sim$18\% speedup.
This is measured against the standard verl recipe, which recomputes old-policy log-probabilities with the training engine because those returned by the rollout engine (e.g., vLLM) differ numerically~\citep{yao2025offpolicy,vllm2025bitwise}; removing this forward pass saves roughly 15\% of per-step compute in our setting.
The gap would be smaller against a baseline that reuses rollout log-probabilities and handles the mismatch (e.g., with truncated importance sampling~\citep{yao2025offpolicy}), whereas LGPO needs no old-policy log-probabilities from either engine.

We also evaluate LGPO in the \emph{fully on-policy} setting ($N_\text{updates}=1$).
For this experiment, we set the minibatch size to 128, matching the prompt batch size, to perform a single gradient update per rollout batch.
As shown in Figure~\ref{fig:fully_on_policy}, GRPO collapses in this setting because the importance ratio is constant at $1$, falling back to a vanilla policy gradient without any trust region constraint.
In contrast, LGPO remains stable (Figure~\ref{fig:fully_on_policy_lgpo_vs_grpo}): its gate depends solely on the current policy's likelihood, so it still filters out high-likelihood samples that induce large policy updates.
Adding a KL penalty toward the reference policy (coefficient $0.001$) does not prevent the collapse of fully on-policy GRPO, whose reward still crashes to zero around step 900, whereas LGPO trains stably without any reference log-probabilities (Appendix~\ref{app:kl_baseline}); a KL penalty constrains the average deviation, but not the per-sample high-sensitivity updates of Theorem~\ref{thm:local-kl-sensitivity}.
The same pattern holds on mathematical reasoning with Qwen2.5-Math-1.5B (Appendix~\ref{app:fully_on_policy_math}).

Having established LGPO's effectiveness and stability, we now turn to a deeper analysis of its underlying mechanisms.

\subsection{Ablation Studies and Analysis}
\label{sec:ablation_studies_and_analysis}

\paragraph{How does LGPO affect generation entropy?}
We observe that LGPO mitigates the rapid entropy collapse typically seen in GRPO during training on both the Countdown task (Figure~\ref{fig:entropy_countdown}) and the mathematical reasoning task (Figure~\ref{fig:entropy_math} in Appendix~\ref{app:entropy_math}).
By gating out updates from already high-likelihood samples, LGPO prevents the policy from becoming overly deterministic, thereby sustaining exploration.
This observation aligns with prior work showing that policy gradient updates on high-probability actions with positive advantages tend to reduce entropy~\citep{cui2025entropy}.
The effect is reminiscent of maximum-entropy RL~\citep{haarnoja2017reinforcement,haarnoja2018soft}, which explicitly rewards policy entropy; LGPO obtains a related effect implicitly, without an entropy bonus, by withholding updates that would further sharpen already near-deterministic responses.

\begin{figure*}[t!]
  \centering
  \begin{minipage}[t]{0.49\textwidth}
    \centering
    \includegraphics[width=\linewidth]{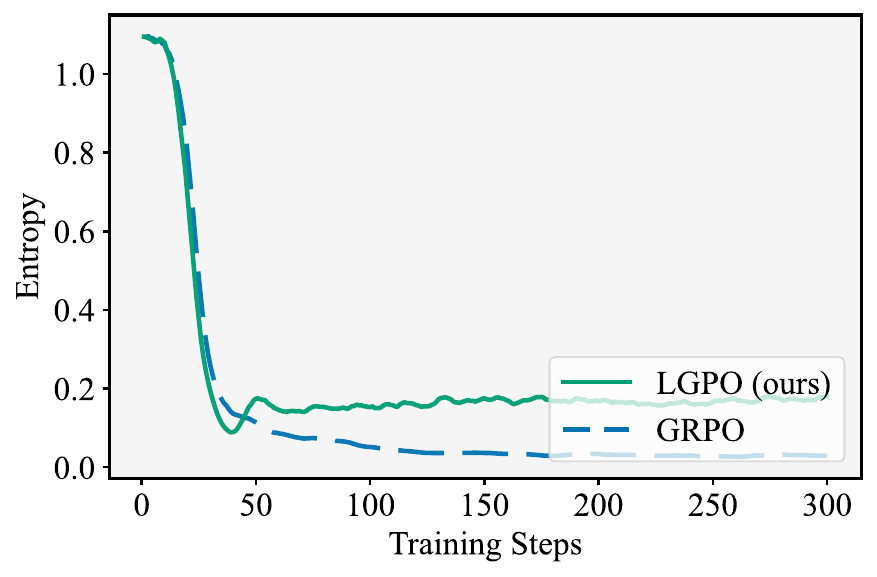}
    \captionof{figure}{Generation entropy during training on Countdown. LGPO maintains higher entropy compared to GRPO, preventing premature convergence to a deterministic policy.}
    \label{fig:entropy_countdown}
  \end{minipage}\hfill
  \begin{minipage}[t]{0.49\textwidth}
    \centering
    \includegraphics[width=\linewidth]{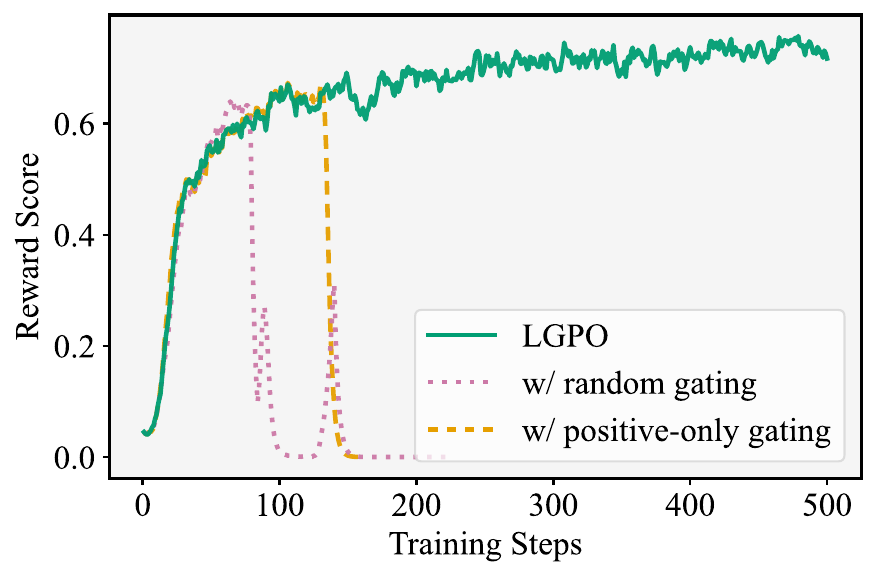}
    \captionof{figure}{Countdown training curves comparing LGPO with random gating regardless of likelihood and gating only high-likelihood samples with positive advantage.}
    \label{fig:random_and_positive_clipping}
  \end{minipage}
\end{figure*}

\paragraph{Do high-likelihood samples really collapse training?}
Randomly discarding samples at LGPO's gating rate, regardless of their likelihood, leads to training collapse within 100 steps (Figure~\ref{fig:random_and_positive_clipping}).
This confirms that LGPO's stability arises from selectively suppressing high-likelihood updates, rather than simply reducing the number of active samples.

\paragraph{Do negative advantage high-likelihood samples also collapse training?}
Negative-advantage updates might seem safer than positive ones, as they push probabilities away from the boundary rather than increasing determinism.
However, as shown in Figure~\ref{fig:random_and_positive_clipping}, gating only positive-advantage high-likelihood samples while keeping negative-advantage ones still leads to collapse.
Suppressing negative-advantage updates on high-likelihood samples is thus also crucial for stability.
We hypothesize two complementary mechanisms.
First, Theorem~\ref{thm:local-kl-sensitivity} is symmetric in the direction of the change, and although raw gradients at saturated tokens are small, Adam makes update magnitudes largely insensitive to gradient scale, so such probabilities can still move, at the highest KL cost per unit of movement.
Second, a sequence-level negative advantage penalizes every token of a failed response, including locally correct tokens the policy is already confident in, which can erode established competencies, consistent with the false-penalty view of \citet{wu2026rethinking}.

\begin{figure*}[t!]
  \centering
  \begin{minipage}[t]{0.51\textwidth}
    \centering
    \includegraphics[width=\linewidth]{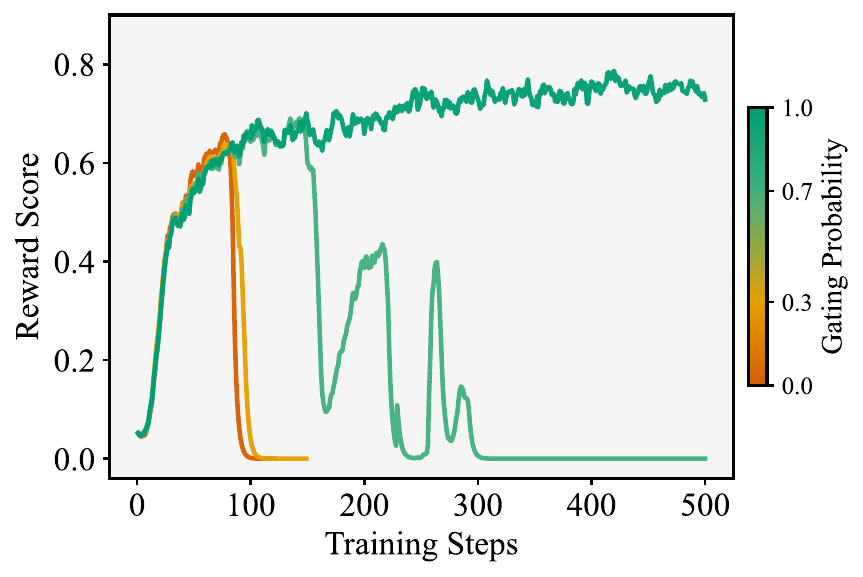}
    \captionof{figure}{Effect of stochastic gating probability on training stability. Reducing the gating probability leads to earlier training collapse, indicating that strict gating is necessary for stability.}
    \label{fig:clipping_chance}
  \end{minipage}\hfill
  \begin{minipage}[t]{0.47\textwidth}
    \centering
    \includegraphics[width=\linewidth]{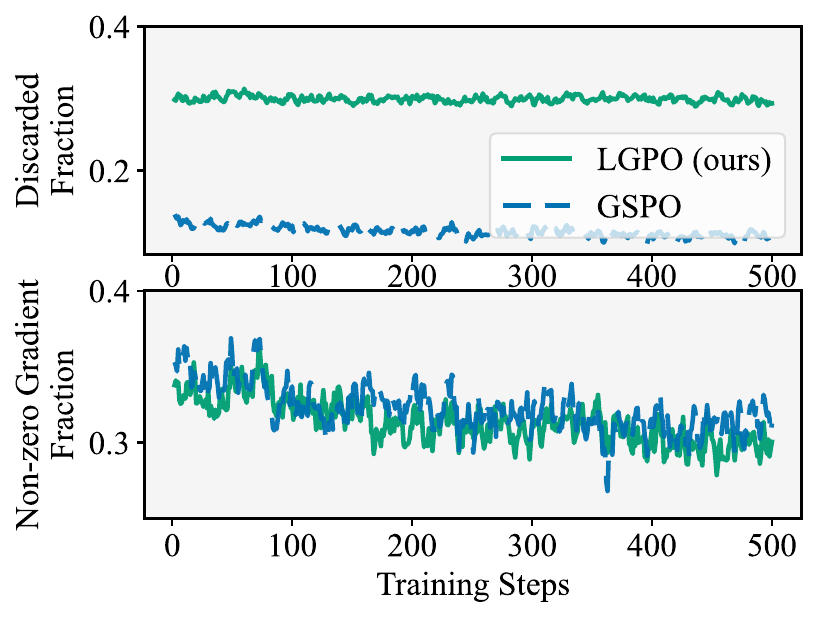}
    \captionof{figure}{Although LGPO discards a larger fraction of total samples (top), the fraction of samples contributing non-zero gradient ($A \neq 0$ and retained) remains comparable to GSPO (bottom).}
    \label{fig:clipping_analysis}
  \end{minipage}
\end{figure*}

\paragraph{Can we stochastically gate high-likelihood samples and find a sweet spot between stability and performance?}
We apply LGPO's gating mask only with a certain probability, to test whether keeping some signal from high-likelihood samples could balance stability and performance.
As shown in Figure~\ref{fig:clipping_chance}, we find no such sweet spot.
Allowing even a small fraction of high-likelihood samples to contribute leads to collapse, with lower gating probabilities causing earlier failure.

\paragraph{Which samples are being gated in LGPO? Are we losing useful training signals?}
To assess how much useful learning signal LGPO discards, we compare it to GSPO, a sequence-level clipping variant of GRPO, as a fair baseline that matches the sequence-level granularity, and track per-step discard statistics throughout training.
Although Figure~\ref{fig:clipping_analysis} (top) shows that LGPO discards a larger fraction of samples overall, we find that a large share of these additionally discarded samples have zero advantage ($A=0$) and would contribute no policy gradient regardless.
Consequently, the effective training signal that remains (measured by the fraction of retained samples with non-zero advantage, i.e., contributing non-zero gradient) stays comparable between LGPO and GSPO across training steps (Figure~\ref{fig:clipping_analysis}, bottom).
With a token-weighted count, LGPO gates 24.7\% of training tokens on Countdown (1.6\% for GSPO), of which 48.9\% have zero, 45.1\% negative, and only 6.1\% positive advantage (Appendix~\ref{app:gated_composition}).
LGPO thus discards little signal from successful trajectories ($\sim$1.5\% of all tokens), but suppresses a substantial share of negative-advantage updates, precisely the category whose retention causes collapse (Figure~\ref{fig:random_and_positive_clipping}).

\section{Related Work}
Reinforcement learning (RL) is increasingly applied to enhance LLM reasoning by adjusting the model's output distribution. Studies suggest that RL updates tend to concentrate probability mass on valid reasoning paths \citep{yue2025does, wang2025beyond}. Additionally, research explores how prolonged training or self-play might extend the solution space beyond the base model's initial capabilities \citep{liu2025prorl, zhao2025absolute}, with recent analysis highlighting the role of entropy regulation in this process \citep{cui2025entropy}.

To address the inherent instability of RL, recent methods focus on refining optimization objectives and learning signals.
One category targets importance sampling variance. Several approaches shift policy optimization from the token level to sequence-level or group-level expectations to reduce volatility \citep{zheng2025group, zhang2025gepo, chen2025minimax}. Others modify constraint mechanisms, such as using selective clipping for negative samples \citep{roux2025tapered}, adaptive gating \citep{gao2025softadaptivepolicyoptimization}, or replacing ratio-based clipping with a direct estimate of policy divergence \citep{qi2026rethinking}, while GVPO alternatively reformulates the objective as a variance-matching regression \citep{zhang2025gvpo}.
The second category improves stability through robust learning signals. This includes enhancing advantage estimation via optimal baselines and global normalization \citep{hao2025policy, xiong2025aapo, xu2025singlestreampolicyoptimization, liu2025understanding}, as well as refining reward aggregation to handle noise \citep{xiao2025bnpo, zhao2025geometricmeanpolicyoptimization}. To prevent reward hacking and collapse, recent works incorporate dynamic sample filtering \citep{lin2025cppo, yu2025dapo} and cross-view invariance constraints \citep{zhang2025co}. \citet{wu2026rethinking} show that negative samples can be unfairly penalized in critic-free RLVR, which relates to our finding that negative-advantage high-likelihood samples destabilize training.

Closest to our analysis, concurrent work~\citep{yao2026grouprelative} shows that group-relative REINFORCE admits an off-policy interpretation and likewise argues that stability stems from regularizing the update rather than from the importance sampling correction.
We complement it with controlled decompositions across four GRPO variants, the structural failure of ratio-based clipping in the fully on-policy regime, and LGPO, a ratio-free trust region that resolves this failure without old-policy likelihoods.
GPG~\citep{chu2026gpg} trains without importance ratios or clipping, which in our setup is stable only with a strong KL penalty.
For strongly off-policy or asynchronous training, TBA~\citep{bartoldson2025trajectory} and OAPL~\citep{ritter2026llms} also dispense with importance ratios, relying instead on KL regularization toward a behavior or reference policy whose log-probabilities must be computed; LGPO targets the fully and near on-policy regime and uses only the current policy's likelihood.
Train--inference mismatch makes nominally on-policy training slightly off-policy~\citep{yao2025offpolicy}; it can be mitigated with truncated importance sampling~\citep{yao2025offpolicy} or bitwise-consistent kernels~\citep{vllm2025bitwise}, whereas LGPO's gate does not use old-policy log-probabilities.

\section{Conclusion}

We propose Likelihood-Gated Policy Optimization (LGPO), motivated by the insight that importance sampling correction is dispensable for stability and by theoretical analysis showing that extreme-likelihood samples induce large policy updates.
By replacing ratio-based clipping with likelihood-based gating, LGPO enforces a trust region without computing old-policy likelihoods, and it remains robust in the fully on-policy setting where ratio-based clipping fails, offering a more efficient and stable framework for LLM RL training.

\section{Limitations and Future Work}

While LGPO demonstrates improved efficiency and stability, our study has limitations.
First, due to computational constraints, our validation covers only 1.5B--3B models on verifiable reasoning tasks; the fully on-policy failure of clipping is structural and Theorem~\ref{thm:local-kl-sensitivity} is model-size agnostic, but transfer to much larger models remains to be verified.
Second, LGPO is stable for every $N_\text{updates}$ we tested (up to 16; Appendix~\ref{app:more_off_policy}), but as the role of the correction grows with off-policyness, we expect it to degrade at sufficiently large $N_\text{updates}$, where KL-regularized approaches such as TBA~\citep{bartoldson2025trajectory} are more appropriate.
Third, our speedup is measured against the standard recipe that recomputes old-policy log-probabilities, and our KL-regularized baseline uses a single coefficient ($0.001$).
Future work includes finer-grained gating within a sequence, combining gating with ratio-based clipping when rollouts are reused (preliminary results in Appendix~\ref{app:more_off_policy} suggest they are complementary), and decoding-stage interventions that avoid generating samples that would be gated anyway.

\begin{ack}
This work was supported by KRAFTON and the University of Wisconsin--Madison, which provided funding and computational resources.
\end{ack}

\bibliographystyle{plainnat}
\bibliography{main}

\appendix

\section{Gradient of the PPO Objective and the Role of Ratio-Based Clipping}
\label{app:gradient_derivation}

In this section, we derive the gradient of the PPO/GRPO surrogate objective to explicitly show how the importance ratio $r_t(\theta)$ acts as the gating mechanism for gradient updates.

Recall the PPO objective from Eq.~\ref{eq:ppo}:
\begin{equation*}
  \mathcal{L}^{\text{PPO}}(\theta) = \mathbb{E} \left[ \sum_{t=1}^{T} \min(r_t(\theta) A_t, \operatorname{clip}(r_t(\theta), 1-\epsilon, 1+\epsilon) A_t) \right],
\end{equation*}
where $r_t(\theta) = \frac{\pi_\theta(y_t|x, y_{<t})}{\pi_{\theta_\text{old}}(y_t|x, y_{<t})}$.

We consider the contribution of a single token at position $t$ to the gradient. Let $L_t(\theta) = \min(r_t(\theta) A_t, \operatorname{clip}(r_t(\theta), 1-\epsilon, 1+\epsilon) A_t)$.
We compute $\nabla_\theta L_t(\theta)$ by analyzing the two cases of the $\min$ operator.

\paragraph{Case 1: $A_t > 0$.}
The objective becomes $\min(r_t(\theta), \operatorname{clip}(r_t(\theta), 1-\epsilon, 1+\epsilon)) A_t$.
The clipping term is upper-bounded by $1+\epsilon$.
\begin{itemize}
    \item If $r_t(\theta) \le 1+\epsilon$, the active term is $r_t(\theta) A_t$. The gradient is $\nabla_\theta (r_t(\theta) A_t) = A_t \nabla_\theta r_t(\theta)$.
    \item If $r_t(\theta) > 1+\epsilon$, the active term is $(1+\epsilon) A_t$. Since this is constant with respect to $\theta$ (locally), the gradient is $\mathbf{0}$.
\end{itemize}

\paragraph{Case 2: $A_t < 0$.}
The objective becomes $\max(r_t(\theta), \operatorname{clip}(r_t(\theta), 1-\epsilon, 1+\epsilon)) A_t$ (taking $A_t$ out effectively flips min to max for the ratio comparison).
\begin{itemize}
    \item If $r_t(\theta) \ge 1-\epsilon$, the active term is $r_t(\theta) A_t$. The gradient is $A_t \nabla_\theta r_t(\theta)$.
    \item If $r_t(\theta) < 1-\epsilon$, the active term is $(1-\epsilon) A_t$. The gradient is $\mathbf{0}$.
\end{itemize}

\paragraph{Gradient Expression.}
Using the identity $\nabla_\theta r_t(\theta) = r_t(\theta) \nabla_\theta \log \pi_\theta(y_t|x, y_{<t})$, we can combine these cases. The gradient is non-zero only when the importance ratio is within the ``active'' unclipped region.
Defining the active indicator $\mathbf{1}_{\text{clip}}$:
\begin{equation*}
\mathbf{1}_{\text{clip}} = \mathbf{1}\left[ (A_t > 0 \land r_t(\theta) \le 1+\epsilon) \lor (A_t < 0 \land r_t(\theta) \ge 1-\epsilon) \right].
\end{equation*}

The gradient for token $t$ is:
\begin{equation*}
\nabla_\theta L_t(\theta) = \mathbf{1}_{\text{clip}} \cdot A_t \cdot r_t(\theta) \cdot \nabla_\theta \log \pi_\theta(y_t|x, y_{<t}).
\end{equation*}

This derivation clarifies that the importance ratio $r_t(\theta)$ serves two roles as discussed in Section \ref{sec:removing_importance_weight}:
(1) scaling the gradient magnitude via the multiplicative term $r_t(\theta)$, and
(2) gating the updates (ratio-based clipping) by zeroing out the gradient via the condition in $\mathbf{1}_{\text{clip}}$. Specifically, if the ratio deviates too far from 1 (exceeds $1+\epsilon$ for positive advantage or drops below $1-\epsilon$ for negative advantage), the gradient is clipped to zero.

\section{Importance Sampling Variance and Bias Analysis}
\label{app:variance_from_large_importance_ratios}

This section explains (1) why large importance ratios can increase the variance of stochastic gradient estimates when used as multiplicative correction weights, and (2) why this specific source of variance disappears when the correction weight is removed (i.e., set to $1$).
We also analyze the bias introduced by removing the importance sampling correction.

\paragraph{Setup and notation.}
Fix a prompt $x$ and consider a response $y=(y_1,\dots,y_{T})$ generated by the old policy,
$y \sim \pi_{\theta_{\text{old}}}(\cdot\mid x)$.
For each token position $t$, define the score function
\[
s_t(\theta) = \nabla_\theta \log \pi_\theta(y_t \mid x, y_{<t}),
\]
and the token-level importance ratio
\[
r_t(\theta) = \frac{\pi_\theta(y_t \mid x, y_{<t})}{\pi_{\theta_{\text{old}}}(y_t \mid x, y_{<t})}.
\]
Let $A_t$ denote the token-level advantage used in the surrogate objective.
We assume absolute continuity: $\pi_{\theta_{\text{old}}}(y_t\mid x,y_{<t})>0$ whenever $\pi_\theta(y_t\mid x,y_{<t})>0$.

\paragraph{Surrogate objective and its gradient.}
Recall the PPO clipped surrogate objective in Eq.~\ref{eq:ppo}.
Without clipping, the resulting surrogate objective for a fixed prompt $x$ can be written as
\[
\mathcal{L}_x(\theta)
:=
\mathbb{E}_{y \sim \pi_{\theta_{\text{old}}}(\cdot\mid x)}\left[\sum_{t=1}^{T} r_t(\theta)\,A_t\right].
\]
Using $\nabla_\theta r_t(\theta) = r_t(\theta)\,\nabla_\theta \log \pi_\theta(y_t\mid x,y_{<t}) = r_t(\theta)\,s_t(\theta)$, the gradient is
\[
\nabla_\theta \mathcal{L}_x(\theta)
=
\mathbb{E}_{y \sim \pi_{\theta_{\text{old}}}(\cdot\mid x)}\left[\sum_{t=1}^{T} r_t(\theta)\,A_t\,s_t(\theta)\right].
\]

\paragraph{Large importance sampling correction can increase variance.}
To analyze variance, we consider the scalar projection of the gradient onto a unit vector $u$:
\[
Z_t^{\text{corr}} := u^\top \big(r_t(\theta)\,A_t\,s_t(\theta)\big) = r_t(\theta)\,A_t\,\big(u^\top s_t(\theta)\big).
\]
Then
\begin{align}
\mathrm{Var}\!\left(Z_t^{\text{corr}}\right)
&=
\mathbb{E}\!\left[\big(Z_t^{\text{corr}}\big)^2\right]
-
\left(\mathbb{E}\!\left[Z_t^{\text{corr}}\right]\right)^2 \nonumber\\
&=
\mathbb{E}\!\left[r_t(\theta)^2\,A_t^2\,\big(u^\top s_t(\theta)\big)^2\right]
-
\left(\mathbb{E}\!\left[r_t(\theta)\,A_t\,\big(u^\top s_t(\theta)\big)\right]\right)^2.
\label{eq:is_var_bound}
\end{align}
Equation~\eqref{eq:is_var_bound} makes the dependence on $r_t(\theta)^2$ explicit through the first term.
If $r_t(\theta)$ is heavy-tailed or occasionally very large, the contribution $\mathbb{E}[r_t(\theta)^2\,A_t^2\,(u^\top s_t(\theta))^2]$ can become large, increasing variance and sensitivity of the gradient estimator.
This is why clipping large $r_t(\theta)$ can reduce the variance contributed by the correction weights.

\paragraph{Variance analysis after removing correction.}
After removing the importance sampling correction by setting the correction weight to $1$, define
\[
Z_t^{\text{no-corr}} := u^\top \big(A_t\,s_t(\theta)\big) = A_t\,\big(u^\top s_t(\theta)\big).
\]
Analogously,
\begin{align*}
\mathrm{Var}\!\left(Z_t^{\text{no-corr}}\right)
&=
\mathbb{E}\!\left[\big(Z_t^{\text{no-corr}}\big)^2\right]
-
\left(\mathbb{E}\!\left[Z_t^{\text{no-corr}}\right]\right)^2 \nonumber\\
&=
\mathbb{E}\!\left[A_t^2\,\big(u^\top s_t(\theta)\big)^2\right]
-
\left(\mathbb{E}\!\left[A_t\,\big(u^\top s_t(\theta)\big)\right]\right)^2,
\end{align*}
which no longer contains $r_t(\theta)$.
Thus, removing correction eliminates the specific variance amplification mechanism caused by large importance ratios.

\paragraph{Bias from removing importance weighting.}
The IS-weighted estimator is unbiased for the on-policy gradient at $\theta$ (assuming integrability), since $\mathbb{E}_{\pi_{\theta_{\text{old}}}}[r_t(\theta)\,A_t\,s_t(\theta)]=\mathbb{E}_{\pi_{\theta}}[A_t\,s_t(\theta)]$ at the token level.
In contrast, the unweighted estimator that sets the correction weight to $1$ targets the behavior-policy expectation and is generally biased.
At the token level, the bias relative to the IS-corrected target is
\begin{align}
\mathrm{Bias}_t(\theta)
&:=
\mathbb{E}_{\pi_{\theta_{\text{old}}}}\!\left[A_t\,s_t(\theta)\right]
-
\mathbb{E}_{\pi_{\theta_{\text{old}}}}\!\left[r_t(\theta)\,A_t\,s_t(\theta)\right] \nonumber\\
&=
\mathbb{E}_{y_t \sim \pi_{\theta_{\text{old}}}(\cdot\mid x,y_{<t})}\!\left[\big(1-r_t(\theta)\big)\,A_t\,s_t(\theta)\right].
\label{eq:nois_bias}
\end{align}
Equation~\eqref{eq:nois_bias} shows that dropping importance weighting trades unbiasedness for potentially reduced variance: the bias vanishes in the fully on-policy setting ($r_t(\theta)\equiv 1$) and grows with the mismatch between $\pi_\theta$ and $\pi_{\theta_{\text{old}}}$, motivating a proximal/trust-region mechanism to keep rollouts from becoming stale across updates.
Note that the clipped PPO/GRPO surrogate is itself already a biased estimator, since clipping discards gradient contributions; our experiments indicate that the additional bias from dropping the correction term is benign for stability and final performance when $N_\text{updates}$ is small (Section~\ref{sec:removing_importance_weight}).

\section{Proof of Theorem~\ref{thm:local-kl-sensitivity}}
\label{app:proof-local-kl-sensitivity}

Let
$p=(p_1,\ldots,p_K)$ denote a categorical distribution over a finite outcome space $\{1,\ldots,K\}$, with $p_i>0$ and $\sum_i p_i=1$.
Let $q=p+\Delta p$ be another distribution, where
$\sum_i \Delta p_i=0$ and $q_i>0$.
The argument below uses only these properties, so it applies to any finite outcome space.
In particular, it covers the next-token distribution $p_i=\pi(i\mid x,y_{<t})$ over a vocabulary (token level) as well as the response distribution $\pi(\cdot\mid x)$ over sequences of bounded length from a finite vocabulary (sequence level), with $q$ induced by an updated policy $\pi'$ in either case.
We write $\mathrm{KL}(p\|q)=\sum_i p_i \log\frac{p_i}{q_i}$.

\begin{lemma}[Quadratic expansion of $\mathrm{KL}$ near $p$]\label{lem:kl-quad-approx}
Let $p_{\min}:=\min_i p_i>0$ and assume $\|\Delta p\|_\infty \le \tfrac12 p_{\min}$.
Then, as $\|\Delta p\|_2\to 0$,
\[
\mathrm{KL}(p\,\|\,p+\Delta p)
= \frac12 \sum_{i=1}^K \frac{(\Delta p_i)^2}{p_i} \;+\; o(\|\Delta p\|_2^2).
\]
\end{lemma}
\begin{proof}
Write $x_i:=\Delta p_i/p_i$, so $q_i=p_i(1+x_i)$ and
\[
\mathrm{KL}(p\|q)=\sum_i p_i\log\frac{p_i}{q_i}=-\sum_i p_i\log(1+x_i).
\]
Using $\log(1+x)=x-\tfrac12 x^2+R(x)$ with $R(x)=O(|x|^3)$ as $x\to 0$, we obtain
\[
\mathrm{KL}(p\|q)
= -\sum_i p_i x_i + \tfrac12\sum_i p_i x_i^2 - \sum_i p_i R(x_i).
\]
Because $\sum_i \Delta p_i=0$, the linear term vanishes: $\sum_i p_i x_i=\sum_i \Delta p_i=0$.
Thus
\[
\mathrm{KL}(p\|q)=\tfrac12\sum_i \frac{(\Delta p_i)^2}{p_i}
- \sum_i p_i R \Big(\tfrac{\Delta p_i}{p_i}\Big).
\]
When $\|\Delta p\|_\infty \le \tfrac12 p_{\min}$ we have $|x_i|\le\tfrac12$ and $|R(x_i)|\le C|x_i|^3$ for a universal $C$,
hence
\[
\Big|\sum_i p_i R\!\Big(\tfrac{\Delta p_i}{p_i}\Big)\Big|
\le C\sum_i \frac{|\Delta p_i|^3}{p_i^2}
\le \frac{C}{p_{\min}^2}\,\|\Delta p\|_\infty \sum_i (\Delta p_i)^2
= O(\|\Delta p\|_2^3)=o(\|\Delta p\|_2^2).
\]
\end{proof}

\begin{lemma}[Minimal quadratic cost for a single coordinate change]\label{lem:kl-min-quad}
Fix $k$ and a small non-zero change $\delta$ with $-p_k < \delta < 1-p_k$.
Among perturbations satisfying $\Delta p_k=\delta$ and $\sum_i \Delta p_i=0$,
the quadratic term in Lemma~\ref{lem:kl-quad-approx} is minimized by
\[
\Delta p_j^\ast = -\,\delta\,\frac{p_j}{1-p_k}\qquad (j\neq k),
\]
and the minimal value equals
\[
\min_{\substack{\Delta p_k=\delta\\ \sum_i \Delta p_i=0}}
\ \frac12\sum_{i=1}^K \frac{(\Delta p_i)^2}{p_i}
= \frac{\delta^2}{2\,p_k(1-p_k)}.
\]
\end{lemma}
\begin{proof}
The constraint $\sum_i \Delta p_i=0$ implies $\sum_{j\neq k}\Delta p_j=-\delta$.
Parameterize $\Delta p_j=-\delta r_j$ for $j\neq k$ with $\sum_{j\neq k} r_j=1$.
Then
\[
\frac12\sum_{i=1}^K \frac{(\Delta p_i)^2}{p_i}
=\frac12\left(\frac{\delta^2}{p_k} + \delta^2\sum_{j\neq k}\frac{r_j^2}{p_j}\right).
\]
Thus it suffices to minimize $\sum_{j\neq k} r_j^2/p_j$ subject to $\sum_{j\neq k} r_j=1$.
Using Lagrange multipliers yields $r_j^\ast \propto p_j$, hence $r_j^\ast=\frac{p_j}{1-p_k}$ and
$\Delta p_j^\ast=-\delta\,\frac{p_j}{1-p_k}$.
Substituting back gives the stated minimum $\frac{\delta^2}{2p_k(1-p_k)}$.
\end{proof}

\begin{proof}[Proof of Theorem~\ref{thm:local-kl-sensitivity}]
By Lemma~\ref{lem:kl-quad-approx},
\[
\mathrm{KL}(p\|p+\Delta p)
= \frac12 \sum_{i=1}^K \frac{(\Delta p_i)^2}{p_i} \;+\; o(\|\Delta p\|_2^2).
\]
Under the constraints $\Delta p_k=\delta$ and $\sum_i \Delta p_i=0$,
Lemma~\ref{lem:kl-min-quad} gives
\[
\min \ \frac12\sum_{i=1}^K \frac{(\Delta p_i)^2}{p_i}
= \frac{\delta^2}{2\,p_k(1-p_k)}.
\]
Moreover, for the minimizing perturbation $\Delta p^\ast$, we have $\|\Delta p^\ast\|_2=\Theta(|\delta|)$ for fixed $p$,
so the remainder term is $o(\|\Delta p^\ast\|_2^2)=o(\delta^2)$.
Combining these yields
\[
\inf_{\substack{\Delta p_k=\delta\\ \sum_i \Delta p_i=0}}
\mathrm{KL}(p\,\|\,p+\Delta p)
= \frac{\delta^2}{2\,p_k(1-p_k)} + o(\delta^2),
\]
as $|\delta|\to 0$, completing the proof.
\end{proof}

\section{Algorithmic and Implementation Details}
\label{app:algorithmic-details}

\begin{algorithm}[H]
\caption{Likelihood-Gated Policy Optimization}
\label{alg:clpo}
\begin{algorithmic}[1]
\REQUIRE current policy $\pi_\theta$, old (behavior) policy $\pi_{\theta_{\text{old}}}$, likelihood threshold $\tau$, number of gradient updates per rollout batch $N_\text{updates}$
\FOR{each rollout batch}
    \STATE Roll out responses $y \sim \pi_{\theta_{\text{old}}}(\cdot\mid x)$ and compute advantages $\{A_t\}_{t=1}^{T}$ to form a rollout buffer $\mathcal{B}=\{(x,y,\{A_t\}_{t=1}^{T})\}$.
    \STATE Partition $\mathcal{B}$ into $N_\text{updates}$ disjoint minibatches $\{\mathcal{B}_n\}_{n=1}^{N_\text{updates}}$.
    \FOR{$n=1$ \textbf{to} $N_\text{updates}$}
        \STATE Initialize minibatch loss $L \gets 0$.
        \FOR{each $(x,y,\{A_t\}_{t=1}^{T})\in \mathcal{B}_n$}
            \STATE $\bar{\ell}_\theta(x,y) \gets \frac{1}{T}\sum_{t=1}^{T}\log \pi_\theta(y_t\mid x,y_{<t})$.
            \STATE $c_\theta(x,y) \gets \mathbf{1}(\bar{\ell}_\theta(x,y) \le \tau)$ \textit{(stop-gradient)}
            \STATE $L \gets L - c_\theta(x,y)\sum_{t=1}^{T} A_t \log \pi_\theta(y_t\mid x,y_{<t})$.
        \ENDFOR
        \STATE Update $\theta$ using $\nabla_\theta L$.
    \ENDFOR
\ENDFOR
\end{algorithmic}
\end{algorithm}

\subsection{Hyperparameters}
\label{app:hyperparameters}

Table~\ref{tab:hyperparameters} lists the training hyperparameters.
All methods share identical hyperparameters within each setting; only the policy objective differs.
Training horizons differ by experiment purpose: 300 steps for the analyses in Section~\ref{sec:removing_importance_weight}, 500 steps for the main comparisons, and 800 steps for fully on-policy runs; the KL-regularized baseline (Appendix~\ref{app:kl_baseline}) is run beyond this horizon.

\begin{table}[h]
  \centering
  \small
  \caption{Training hyperparameters.}
  \label{tab:hyperparameters}
  \begin{tabular}{lcc}
    \toprule
    Hyperparameter & Countdown & Mathematical reasoning \\
    \midrule
    \multirow{2}{*}{Models} & \multirow{2}{*}{Qwen2.5-3B-Instruct} & Qwen2.5-3B-Instruct \\
     & & Qwen2.5-Math-1.5B \\
    Framework & \multicolumn{2}{c}{verl~\citep{sheng2025hybridflow}} \\
    Optimizer & \multicolumn{2}{c}{AdamW} \\
    Learning rate & \multicolumn{2}{c}{$1\times 10^{-6}$} \\
    LR warmup steps & \multicolumn{2}{c}{10} \\
    Weight decay & \multicolumn{2}{c}{0.1} \\
    Prompt batch size & 128 & 256 \\
    Minibatch size & 64 & 64 \\
    $N_\text{updates}$ & 2 & 4 \\
    Minibatch size, fully on-policy ($N_\text{updates}=1$) & 128 & -- \\
    Group size $G$ & 8 & 8 \\
    Advantage normalization & \multicolumn{2}{c}{group mean and std.\ (GRPO)} \\
    KL penalty & \multicolumn{2}{c}{none (except KL baseline)} \\
    LGPO threshold $\tau$ & \multicolumn{2}{c}{$-0.15$} \\
    Evaluation & -- & average Pass@1 over 8 samples \\
    \bottomrule
  \end{tabular}
\end{table}

\section{Additional Experiments Results}
\label{app:additional_experiments_results}

\subsection{Removing Importance Sampling Correction in DAPO and GSPO}
\label{app:dapo_gspo_removing_correction}

\begin{figure}[ht]
  \centering
  \begin{subfigure}[b]{0.48\textwidth}
    \centering
    \includegraphics[width=\linewidth]{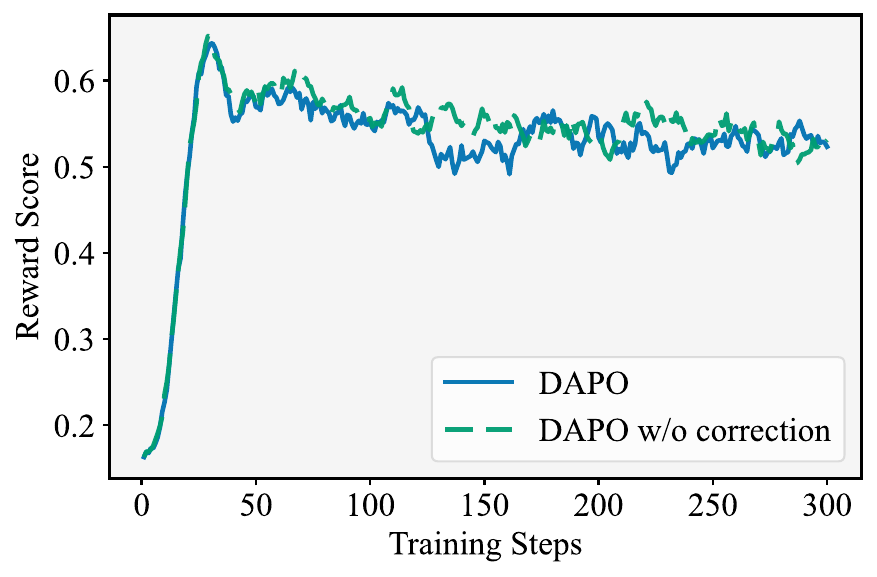}
    \caption{DAPO}
    \label{fig:dapo_removing_importance_weight}
  \end{subfigure}
  \hfill
  \begin{subfigure}[b]{0.48\textwidth}
    \centering
    \includegraphics[width=\linewidth]{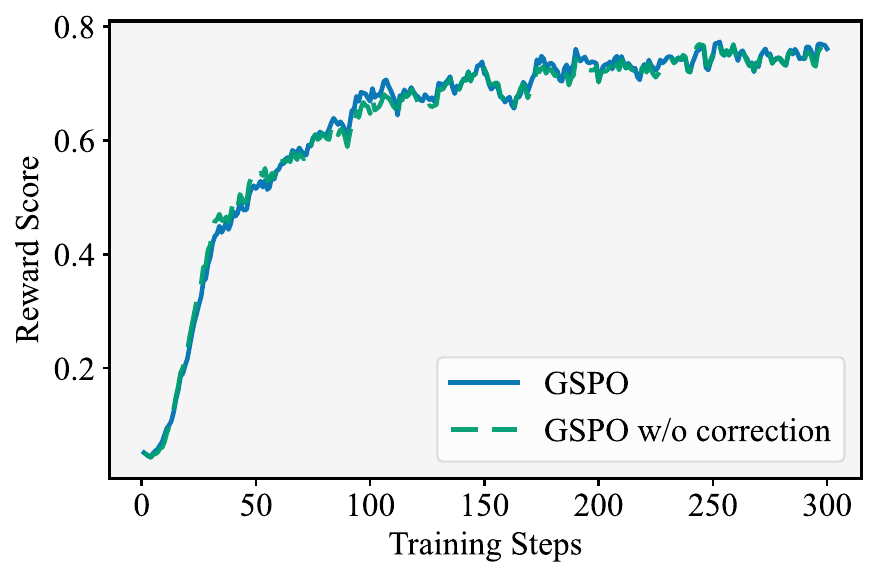}
    \caption{GSPO}
    \label{fig:gspo_removing_importance_weight}
  \end{subfigure}
  \caption{The reward score of the Countdown task during the RL training. Removing the importance sampling correction has little effect on training stability for both DAPO and GSPO.}
  \label{fig:dapo_gspo_removing_correction}
\end{figure}

In Section~\ref{sec:removing_importance_weight}, we show that removing the importance sampling correction from GRPO has little effect on training stability.
We further investigate the effect of removing the importance sampling correction in other GRPO-variants, specifically DAPO and GSPO.
Similar to our findings with GRPO, removing the correction term alone does not destabilize training for these algorithms (Figure~\ref{fig:dapo_gspo_removing_correction}).

\subsection{Likelihood Distribution of Sampled Responses}
\label{app:likelihood_distribution}

\begin{figure}[h]
  \centering
  \includegraphics[width=0.5\linewidth]{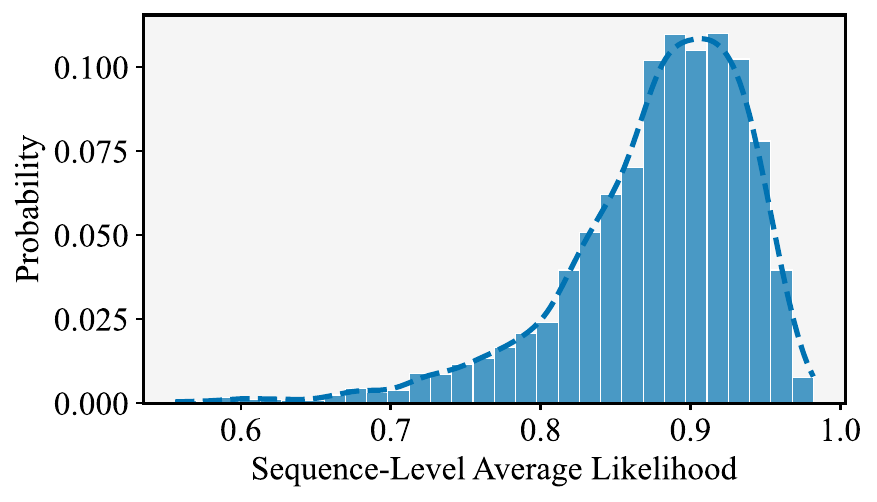}
  \caption{Distribution of sequence-level average likelihood $\exp \left( \bar{\ell}_\theta(x,y) \right)$ of responses sampled from Qwen2.5-3B-Instruct on MATH500.}
  \label{fig:likelihood_distribution}
\end{figure}

Figure~\ref{fig:likelihood_distribution} shows the distribution of the sequence-level average likelihood of responses sampled from Qwen2.5-3B-Instruct on the MATH500 dataset.
Since low-likelihood sequences are rarely observed, LGPO only gates high-likelihood sequences to enforce the trust region constraint.

\subsection{Generation Entropy on Mathematical Reasoning Task}
\label{app:entropy_math}

\begin{figure}[ht]
  \centering
  \includegraphics[width=0.5\linewidth]{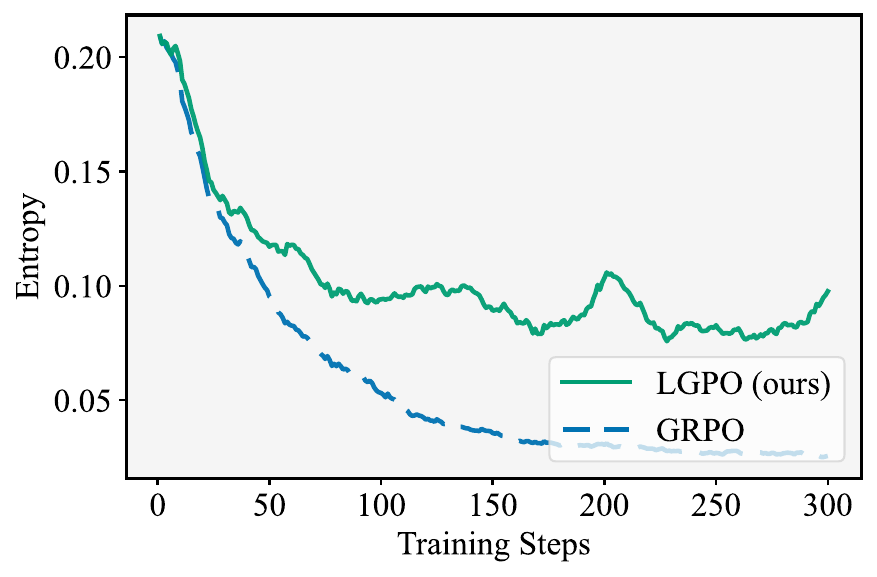}
  \caption{Generation entropy during training on the mathematical reasoning dataset. LGPO prevents rapid entropy collapse, similar to the trend observed in the Countdown task.}
  \label{fig:entropy_math}
\end{figure}

Figure~\ref{fig:entropy_math} presents the generation entropy during training on the mathematical reasoning dataset.
Similar to the results on the Countdown task, LGPO maintains higher entropy throughout training compared to GRPO, which exhibits a more rapid decline in entropy.

\section{Additional Analyses and Baselines}
\label{app:discussion_results}

This section reports additional analyses and baselines.
Unless otherwise noted, experiments use Qwen2.5-3B-Instruct on Countdown with the setup and hyperparameters of Section~\ref{sec:experiments} and Appendix~\ref{app:hyperparameters}.

\subsection{Decomposing the Importance Ratio in Dr.~GRPO}
\label{app:dr_grpo}

Dr.~GRPO~\citep{liu2025understanding} computes advantages without dividing by the group standard deviation, as is common in current GRPO implementations.
We repeat the decomposition of Section~\ref{sec:removing_importance_weight} with Dr.~GRPO-style advantages on Countdown, matching the setup of Figure~\ref{fig:removing_importance_weight} (300 steps).
As shown in Table~\ref{tab:dr_grpo}, removing the correction term alone does not harm stability, so together with GRPO, DAPO, and GSPO (Figure~\ref{fig:dapo_gspo_removing_correction}), the dispensability of the correction holds across four GRPO variants.
Unlike for GRPO, removing clipping did not destabilize Dr.~GRPO within this horizon.
We attribute this to update magnitude: dividing by the group standard deviation inflates advantages on low-variance groups, so standard GRPO takes larger per-sample updates and depends more heavily on a trust region; the need for a trust region thus scales with update magnitude.
Note that Dr.~GRPO without clipping still computes importance ratios and the old-policy forward pass, so it provides none of LGPO's efficiency benefit.
Whether it remains stable over longer horizons, on other tasks, or in the fully on-policy regime is left open.

\begin{table}[h]
  \centering
  \small
  \caption{Final reward on Countdown after 300 steps when decomposing the importance ratio in Dr.~GRPO.}
  \label{tab:dr_grpo}
  \begin{tabular}{lc}
    \toprule
    Variant & Final reward \\
    \midrule
    Dr.~GRPO & 0.74 \\
    Dr.~GRPO \textit{w/o correction} & 0.71 \\
    Dr.~GRPO \textit{w/o clipping} & 0.73 \\
    \bottomrule
  \end{tabular}
\end{table}

\subsection{GPG-Style Objective without Importance Ratio and Clipping}
\label{app:gpg}

GPG~\citep{chu2026gpg} reports that RL with verifiable rewards can be trained successfully without either an importance ratio or clipping.
We reproduce a GPG-style objective under our controlled Countdown setup, with data, model, and hyperparameters identical to Figure~\ref{fig:removing_importance_weight}.
Without an importance ratio or clipping, it collapses within approximately 120--180 steps in both of our runs, consistent with our \textit{GRPO w/o corr.\ \& clip.}\ results (Figure~\ref{fig:main_training_curves}).
It can be stabilized by adding a sufficiently strong KL loss, i.e., by reintroducing a trust-region-like mechanism.
GPG's reported successes therefore show that training without clipping or an importance ratio can be stable under some configurations, while our results show that it is not uniformly stable across settings.

\subsection{KL-Regularized GRPO in the Fully On-Policy Setting}
\label{app:kl_baseline}

A natural alternative for stabilizing fully on-policy training is a KL penalty toward the reference policy.
We train fully on-policy GRPO ($N_\text{updates}=1$) on Countdown with a KL penalty of coefficient $0.001$, extending this run beyond the 800-step horizon of the other fully on-policy runs.
The KL penalty does not prevent collapse: the reward crashes to zero around step 900, whereas LGPO trains smoothly throughout with a higher final reward, without computing any reference log-probabilities.
We believe this is because a KL penalty constrains the average deviation from the reference policy, but does not suppress the per-sample high-sensitivity updates identified by Theorem~\ref{thm:local-kl-sensitivity}, which LGPO's gating removes.
We tested only this coefficient; a stronger KL penalty may delay or prevent collapse, as it does for the GPG-style objective (Appendix~\ref{app:gpg}), at the cost of restricting deviation from the initialization and of computing reference log-probabilities.

\subsection{More Off-Policy Training}
\label{app:more_off_policy}

To assess robustness to more off-policy data, we train GRPO and LGPO on Countdown with $N_\text{updates}=8$ and $N_\text{updates}=16$ (Table~\ref{tab:more_off_policy}).
All runs train stably over the reported horizons, and LGPO reaches a higher peak and final reward than GRPO in both settings.
At $N_\text{updates}=16$, we additionally run a preliminary combination of the two mechanisms, applying the sequence-level likelihood gate on top of the clipped importance-ratio objective (\textit{LGPO + clipping}).
Without any tuning, the combination improves over clipping alone in both peak and final reward and moderates the entropy inflation that clipping alone shows at this reuse depth, indicating that the gate removes harmful updates that the ratio-based trust region does not filter.
Gating alone is sufficient for both stability and the best performance at $N_\text{updates}=16$, although this offers no guarantee in more strongly off-policy regimes: since the role of the importance sampling correction grows with off-policyness, we expect LGPO to degrade at sufficiently large $N_\text{updates}$, where KL-regularized approaches designed for strongly off-policy training~\citep{bartoldson2025trajectory,ritter2026llms} are more appropriate.

\begin{table}[h]
  \centering
  \small
  \caption{Countdown training with more off-policy data. Final reward is reported at the last step (400 steps for $N_\text{updates}=8$, 800 steps for $N_\text{updates}=16$). All runs are stable.}
  \label{tab:more_off_policy}
  \begin{tabular}{lcccc}
    \toprule
    Method & $N_\text{updates}$ & Peak reward & Final reward & Final entropy \\
    \midrule
    GRPO & 8 & 0.74 & 0.72 & -- \\
    LGPO & 8 & 0.77 & 0.76 & -- \\
    \midrule
    GRPO (clipping only) & 16 & 0.68 & 0.66 & 2.21 \\
    LGPO (gating only) & 16 & 0.78 & 0.76 & 0.80 \\
    LGPO + clipping & 16 & 0.70 & 0.70 & 1.64 \\
    \bottomrule
  \end{tabular}
\end{table}

\subsection{Two-Sided Likelihood Gating}
\label{app:two_sided}

Theorem~\ref{thm:local-kl-sensitivity} is two-sided, whereas LGPO gates only high-likelihood sequences.
To verify directly that gating the low-likelihood side is unnecessary, rather than inferring it from the rarity of low-likelihood sequences alone (Figure~\ref{fig:likelihood_distribution}), we additionally gate sequences whose average likelihood is below $0.6$ (i.e., $\bar{\ell}_\theta(x,y) < \log 0.6$), the lower edge of the distribution in Figure~\ref{fig:likelihood_distribution}, on top of the high-likelihood gate $\bar{\ell}_\theta(x,y) > \tau$.
Two-sided gating trains stably with reward comparable to one-sided LGPO ($0.71$ vs.\ $0.72$ at 500 steps), so gating the low-likelihood side provides no additional stability or performance benefit.
The two sides are not symmetric for a sequence-level statistic: a high average log-likelihood forces most tokens of the sequence toward saturation ($p_k\to 1$), the sensitive regime of Theorem~\ref{thm:local-kl-sensitivity}, whereas a low average log-likelihood instead reflects many moderate-probability tokens, for which $p_k(1-p_k)$ is large and the sensitivity $\delta^2/(2p_k(1-p_k))$ is small; tokens with probability truly near zero are rarely sampled in on-policy data in the first place.

\subsection{Sensitivity to the Likelihood Threshold}
\label{app:tau_sensitivity}

We sweep the likelihood threshold $\tau$ on Countdown.
Our default $\tau=-0.15$ is among the settings that give the most stable training.
Looser thresholds (closer to $0$), which gate only sequences with very high average likelihood, admit destabilizing high-likelihood samples and collapse early, whereas tighter thresholds (more negative) discard too much learning signal and plateau at a lower reward before destabilizing.
The same $\tau=-0.15$ is used unchanged across both models (Qwen2.5-3B-Instruct and Qwen2.5-Math-1.5B) and both tasks.
In practice, we recommend setting $\tau$ near the top 10--20\% quantile of the sequence-level average likelihood of sampled responses (cf.\ Figure~\ref{fig:likelihood_distribution}).

\subsection{Fully On-Policy Training on Mathematical Reasoning}
\label{app:fully_on_policy_math}

To check that the fully on-policy findings are not specific to one model and task, we also train Qwen2.5-Math-1.5B on the mathematical reasoning dataset with $N_\text{updates}=1$.
As on Countdown, fully on-policy GRPO degrades sharply late in training, while LGPO remains stable.

\subsection{Composition of Gated Tokens}
\label{app:gated_composition}

Table~\ref{tab:gated_composition} breaks down, with a token-weighted count over the Countdown training run, which tokens LGPO gates.
LGPO gates 24.7\% of all training tokens, compared with 1.6\% for GSPO's clipping.
Zero-advantage tokens come from groups in which all sampled responses receive the same binary reward and therefore contribute no policy-gradient signal regardless of gating.
Positive-advantage tokens gated by LGPO amount to only 1.5\% of all training tokens, so LGPO discards little signal from successful trajectories; the largest non-zero category it removes is negative-advantage, high-likelihood samples, whose retention leads to collapse (Figure~\ref{fig:random_and_positive_clipping}).

\begin{table}[h]
  \centering
  \small
  \caption{Token-weighted composition of tokens gated by LGPO on Countdown, by advantage sign.}
  \label{tab:gated_composition}
  \begin{tabular}{lcc}
    \toprule
    Advantage of gated tokens & Share of gated tokens & Share of all training tokens \\
    \midrule
    Zero ($A=0$) & 48.9\% & 12.1\% \\
    Negative ($A<0$) & 45.1\% & 11.1\% \\
    Positive ($A>0$) & 6.1\% & 1.5\% \\
    \midrule
    Total gated by LGPO & 100\% & 24.7\% \\
    Total clipped by GSPO (reference) & -- & 1.6\% \\
    \bottomrule
  \end{tabular}
\end{table}

\end{document}